\documentclass{article}

\PassOptionsToPackage{numbers, compress}{natbib}

\usepackage{graphicx}
\usepackage{subcaption}

    \usepackage[final]{neurips_2021}

\usepackage[utf8]{inputenc} 
\usepackage[T1]{fontenc}    
\usepackage{hyperref}       
\usepackage{url}            
\usepackage{booktabs}       
\usepackage{amsfonts}      
\usepackage{nicefrac}       
\usepackage{microtype}      
\usepackage{xcolor}        

\title{Tradeoffs Between Contrastive and Supervised Learning: An Empirical Study}

\author{
  Ananya Karthik, Mike Wu, Noah Goodman, Alex Tamkin \\
  Department of Computer Science\\
  Stanford University\\
  \texttt{\{ananya23,wumike,ngoodman,atamkin\}@stanford.edu} \\
}

\begin{document}

\maketitle

\begin{abstract}

Contrastive learning has made considerable progress in computer vision, outperforming supervised pretraining on a range of downstream datasets. However, is contrastive learning the better choice in all situations? We demonstrate two cases where it is not. First, under sufficiently small pretraining budgets, supervised pretraining on ImageNet consistently outperforms a comparable contrastive model on eight diverse image classification datasets. This suggests that the common practice of comparing pretraining approaches at hundreds or thousands of epochs may not produce actionable insights for those with more limited compute budgets. Second, even with larger pretraining budgets we identify tasks where supervised learning prevails, perhaps because the object-centric bias of supervised pretraining makes the model more resilient to common corruptions and spurious foreground-background correlations. These results underscore the need to characterize tradeoffs of different pretraining objectives across a wider range of contexts and training regimes.

\end{abstract}

\begin{figure}[ht]
\vskip 0.2in
\begin{center}
\centerline{\includegraphics[width=\columnwidth]{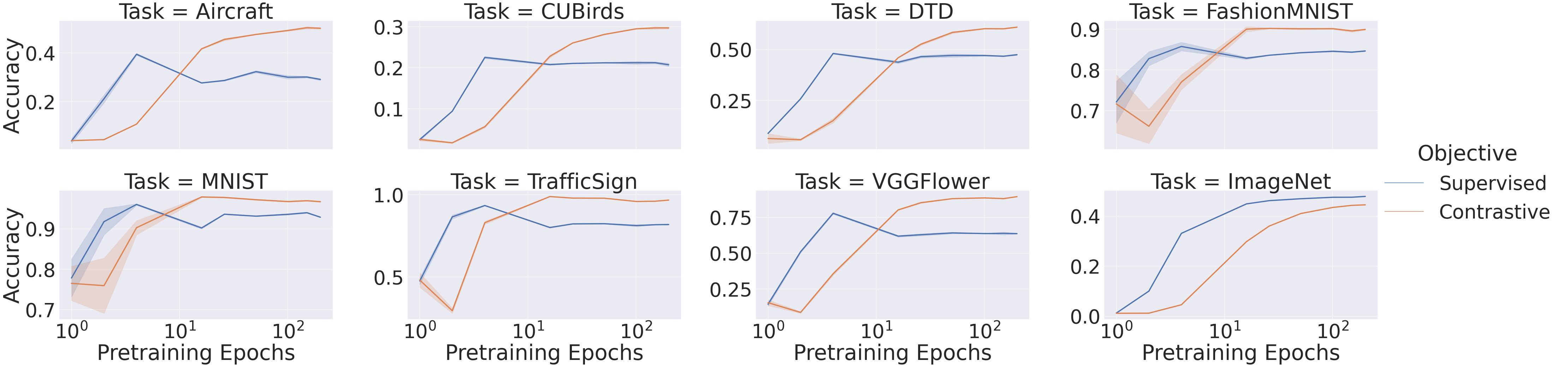}}
\caption{\textbf{Downstream accuracy of contrastive and supervised models for different pretraining budgets.} Models pretrained on ImageNet, then evaluated on 8 diverse image classification datasets. Shaded regions show the standard deviation across three runs (often too small to see without magnification; shown for all except ImageNet which had only one trial). Unpretrained models shown on far left of each plot.}
\label{fig:dynamics}
\end{center}
\vskip -0.2in
\end{figure}

\section{Introduction}

The cost of labeling large-scale datasets has motivated a rise in self-supervised pretraining, with recent methods in computer vision closing the gap with or even surpassing supervised approaches \citep{Caron2020UnsupervisedLO,Grill2020BootstrapYO,He2020MomentumCF,Chen2020ASF,zbontar2021barlow}. Instead of using labels, recent contrastive learning methods leverage an instance discrimination task  \citep{Wu2018UnsupervisedFL,Dosovitskiy2016DiscriminativeUF,Misra2020SelfSupervisedLO, Chen2020ASF,Purushwalkam2020DemystifyingCS} to achieve state-of-the-art results on a variety of computer vision tasks. The instance discrimination task treats each image as its own class, training a model to determine whether two augmented examples were derived from the same original instance using a contrastive loss \citep{Gutmann2010NoisecontrastiveEA,Hnaff2020DataEfficientIR,Tian2020ContrastiveMC}. This training procedure produces models whose representations are broadly useful for a range of transfer tasks \citep{Ericsson2020HowWD,Yang2020TransferLO}.

These advances underscore the need for studying the real-world tradeoffs between contrastive and supervised pretraining. We investigate by comparing the transfer performance of both methods across different pretraining budgets and transfer datasets. Our results address two questions:
\begin{enumerate}
    \item \textbf{Is contrastive learning better than supervised across all compute budgets?} No, different pretraining algorithms produce better representations at different pretraining budgets. Moreover, transfer accuracy on different tasks is not even monotonic across pretraining. Thus, we recommend that future work on pretraining report transfer accuracy across epochs so practitioners can make informed decisions based on their end task and compute budget.
    \item \textbf{For larger compute budgets, is contrastive pretraining better for all tasks?} No. While the supervised model eventually achieves worse downstream accuracy than the contrastive model on most tasks, we identify tasks where the object-centric bias of ImageNet pretraining aids transfer---especially in the Waterbirds dataset, which measures reliance on spurious correlations and ImageNet-C, which measures robustness to common corruptions. 
\end{enumerate}

\section{Related Work}

\paragraph{Performance of self-supervised learning} Previous studies on representation learning for visual tasks have provided insights into the generalizability and transfer performance of various algorithms, including the comparison of supervised and unsupervised learning methods \citep{cole2021does,kotar2021contrasting,Kornblith2019DoBI,Goyal2019ScalingAB,Kolesnikov2019RevisitingSV,Zhai2019ALS,Ericsson2020HowWD,Yang2020TransferLO}. 
In particular, \citet{Ericsson2020HowWD} find that the best self-supervised models outperform a supervised baseline on most datasets in their benchmark. Our work builds on this analysis by holding variables like pretraining epochs and data augmentations constant, performing a controlled analysis of these models' learning dynamics and transferability.

\paragraph{Sample efficiency of pretraining methods} Several studies \citep{Zhai2019ALS,Goyal2019ScalingAB,Kolesnikov2019RevisitingSV,Yang2020TransferLO} analyze sample efficiency or computational efficiency during transfer or finetuning, including \citet{Zhao2020WhatMI}, which compares the pretraining dynamics of supervised and unsupervised learning methods on the VOC'07 detection task. However, to the best of our knowledge, there has not been a comprehensive comparison of the learning dynamics over the course of pretraining time for both supervised and contrastive learning across a diverse range of downstream image classification tasks.

\paragraph{How pretraining objectives shape model representations} \citet{Zhao2020WhatMI} visualize the representations of contrastive and supervised models, arguing that the former objective may produce more holistic representations compared to the latter. Furthermore, \citet{Ericsson2020HowWD} observe that self-supervised pretraining attends to larger regions than supervised pretraining, a characteristic that may aid the transfer performance of self-supervised methods. \citet{cole2021does} and \citet{vanhorn2021benchmarking} show that self-supervised pretraining does not outperform supervised learning for fine-grained classification tasks. Our study builds upon these works by providing examples of specific object-centric \textit{tasks} where the supervised model achieves higher accuracy, as well as cases where the holistic representations of the contrastive model prevail. 

\section{Experiments}

\subsection{Experimental Settings}

We pretrain two ResNet-18 models on ILSVRC-2012 (ImageNet) \citep{Russakovsky2015ImageNetLS} for 200 epochs with a batch size of 128. We use the standard cross entropy loss for the supervised model, and we use the InfoNCE  objective from \citet{Wu2018UnsupervisedFL} for the contrastive model, leveraging a memory bank for negatives. Both models are pretrained with identical image augmentations, the same as \citet{Chen2020ASF} without random Gaussian blur, and identical model architecture. For pretraining, we use SGD with a learning rate of 0.03, momentum of 0.9, and weight decay of 1e-4.

For transfer, we use the linear evaluation protocol \citep{Chen2020ASF}, training a logistic regression model on the outputs of the prepool 512x7x7 layer of a frozen pretrained model. We evaluate both pretrained models by training them for 100 epochs on eight transfer tasks: MNIST \citep{LeCun1998GradientbasedLA}, FashionMNIST \citep{Xiao2017FashionMNISTAN}, VGG Flower (VGGFlower) \citep{Nilsback2008AutomatedFC}, Traffic Signs (TrafficSign) \citep{Houben2013DetectionOT}, Describable Textures (DTD) \citep{Cimpoi2014DescribingTI}, CUB-200-2011 (CUBirds) \citep{Wah2011TheCB}, Aircraft \citep{Maji2013FineGrainedVC}, and ImageNet itself. (See Table \ref{fig:datasets} in the Appendix.) We use SGD with a batch size of 256, learning rate of 0.01, momentum of 0.9, and weight decay of 1e-4.

\subsection{Results}
We first compare final transfer accuracies achieved by contrastive and supervised pretraining (Figure \ref{fig:transfer}). In line with previous studies \citep{Ericsson2020HowWD,Chen2020ASF}, we find that the contrastive model outperforms the supervised model on all transfer tasks except ImageNet, its pretraining dataset.

\subsubsection{Learning Dynamics Across Compute Budgets}

We also investigate the representation learning dynamics and computational efficiency of the two models. Transfer accuracy by pretraining budget is shown in Figure \ref{fig:dynamics}. All results except ImageNet represent the average of three trials with different random seeds; ImageNet results represent one trial. We observe the following trends across the seven non-ImageNet tasks:

\textbf{(i)} With only a few epochs of pretraining, the supervised model maintains a lead over the contrastive model. However, by 15 epochs the contrastive model rapidly overtakes the supervised model's downstream accuracy, maintaining a lead until the end of pretraining. Thus, the contrastive model is more computationally efficient for all but the most restricted compute budgets. However, it also suggests a note of caution: models that prevail after a certain number of pretraining steps may not always win out at other, more modest budgets.

\textbf{(ii)} The downstream accuracy of both models does not always increase monotonically across pretraining. Particularly pronounced for the supervised model, this phenomenon suggests a potential misalignment between the representations developed for the supervised task and those most useful for the downstream tasks.

\subsubsection{Downstream Effects of Biases Acquired During Pretraining}
\label{sec:object-centric}

We have demonstrated that the supervised and contrastive models have different pretraining dynamics, suggesting that the models may acquire different feature processing capabilities during pretraining. But what are the downstream effects of these representational differences? \citet{Zhao2020WhatMI} conclude that supervised pretraining may learn more object-specific features than contrastive models. In three controlled studies (see Figure \ref{fig:DatasetExamples} in the Appendix for example images from the datasets used), we investigate this hypothesis by examining specific tasks where an object-centric bias may be salient. All results are averages of three trials with different random seeds.

\paragraph{NORB} We study the transfer performance of both models on a carefully-controlled dataset which isolates the models' abilities to capture both object and non-object information in their representations. Specifically, we use NORB (small set\footnote{\url{https://cs.nyu.edu/\~ylclab/data/norb-v1.0-small/}}) --- synthetic images of 50 types of toys, annotated with toy category, lighting conditions, elevations, and azimuths \citep{LeCun2004LearningMF}. Contrastive pretraining outperforms supervised learning on object, elevation, lighting, and azimuth classification tasks. For the non-object elevation and lighting transformations, the gap in accuracy between the models was pronounced --- 16.18\% and 23.91\%, respectively --- possibly due to the supervised model developing more object-centric representations. However, drawing firm conclusions is challenging, as the accuracy difference across tasks may be misleading when object classification accuracy approaches the 100\% ceiling. Furthermore, even if the gap differs, the contrastive model still outperforms the supervised model across tasks. Thus, these results provide relatively modest evidence of transfer tasks where object-centricity impacts the two models differently.

\begin{figure}[h]
\vskip 0pt
\begin{subfigure}{.48\textwidth}
    \centering
\includegraphics[width=\columnwidth]{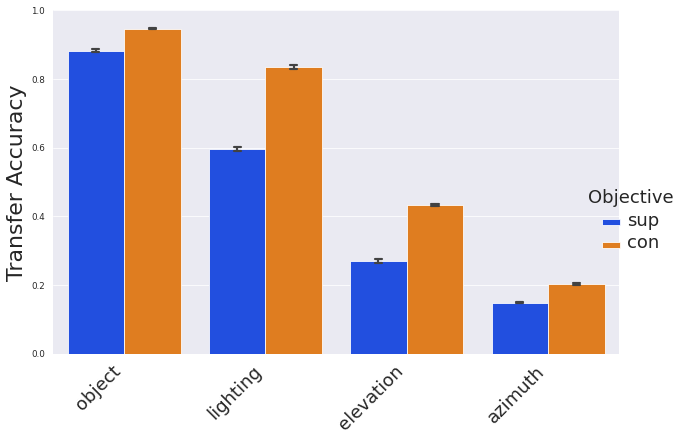}
    \caption{\textbf{Transfer accuracy on NORB object, elevation, lighting, and azimuth classification.} Contrastive accuracy was higher than supervised accuracy on all tasks. Models pretrained on ImageNet for 200 epochs; average of three trials, with error bars for standard deviations.}
    \label{fig:norb}
\end{subfigure}
\hfill
\begin{subfigure}{.48\textwidth}
    \centering
\includegraphics[width=\columnwidth]{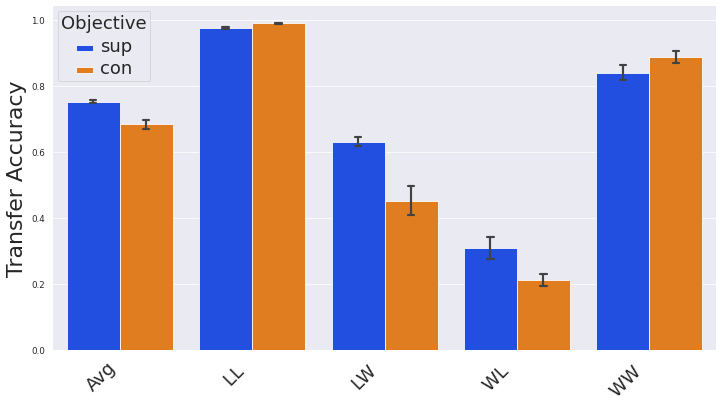}
    \caption{\textbf{On the Waterbirds dataset, the supervised model appears to attend less to spurious correlations.} \textbf{LW} indicates images of \textbf{L}and birds on \textbf{W}ater backgrounds. Models pretrained on ImageNet for 200 epochs; average of three trials, with error bars for standard deviations.}
    \label{fig:waterbirds}
\end{subfigure}
\caption{\textbf{Difference in learned representations: NORB and Waterbirds.}}
\label{fig:norb+waterbirds}
\end{figure}

\paragraph{Waterbirds} We then expand our experiments to a different set of non-object properties --- the content of image backgrounds. We evaluate contrastive and supervised pretraining on Waterbirds, a dataset designed to examine spurious correlations based on the relationship between object and background (see Appendix) \citep{Sagawa2019DistributionallyRN}. We find that on images in which the backgrounds and objects are mismatched the supervised model achieves higher transfer accuracy than the contrastive model, in contrast to the previous results showing higher image classification performance for the contrastive model. This suggests that the supervised model may have learned more object- or foreground-centric representations, which render the spurious background feature less prominent. While this result lends support to the notion that the contrastive model learns a more holistic image representation, it also suggests that the inductive bias attained from a more tailored representation may be helpful in underspecified settings where undesired features are also predictive of the class of interest \citep{d2020underspecification}.

\paragraph{ImageNet-C} Last, we study the degradation of transfer performance in the presence of non-object-based corruptions. We hypothesized that if supervised learning results in more object-centric representations, then transfer performance might degrade less with non-object corruptions such as color shifts and changes in contrast. We evaluate both models, after transfer was performed for ImageNet, on 15 corruptions from ImageNet-C, a dataset created by applying 15 corruptions at 5 severity levels to ImageNet validation images \citep{Hendrycks2019BenchmarkingNN}. We observe the relative mCE, which measures the performance degradation from clean to corrupted data  (lower is better), to be lower for supervised ($\textbf{91.08} \pm \textbf{0.279\%}$) vs the contrastive model ($\textbf{95.41} \pm \textbf{0.157\%}$). This provides additional evidence that supervised pretraining may lead to more object-centric representations than contrastive approaches.

\section{Discussion}

We investigate tradeoffs between supervised and contrastive pretraining. 

Our first set of experiments examines how the linear evaluation performance on a range of transfer tasks changes as each model pretrains. Surprisingly, we find that transfer performance does not monotonically increase across pretraining, suggesting a misalignment between representations learned for pretraining vs transfer. Moreover, while the contrastive model eventually achieves higher performance, for the first 10-15 epochs the supervised model yields better representations for downstream tasks. This not only reveals differences in the process by which both models acquire their useful representations, but also that conclusions drawn for models trained for thousands of epochs may not always transfer over to practitioners with more modest compute budgets. Thus, we encourage developers of new pretraining techniques to release learning dynamics curves so that practitioners can make decisions based on their own budgets and use cases.

To further explore tradeoffs between the two models, we examine whether supervised learning imparts an object-centric bias detectable through improved performance on transfer tasks. We find strong effects in the case of Waterbirds and ImageNet-C, but weaker effects for the NORB dataset. We encourage future work investigating how pretraining objectives shape the behavior of models in ambiguous scenarios, as well as more broadly investigating whether these conclusions hold across a wider range of architectures, hyperparameters, datasets, and training objectives.

\bibliographystyle{plainnat}
\bibliography{neurips_2021_bib}

\begin{thebibliography}{34}
\providecommand{\natexlab}[1]{#1}
\providecommand{\url}[1]{\texttt{#1}}
\expandafter\ifx\csname urlstyle\endcsname\relax
  \providecommand{\doi}[1]{doi: #1}\else
  \providecommand{\doi}{doi: \begingroup \urlstyle{rm}\Url}\fi

\bibitem[Caron et~al.(2020)Caron, Misra, Mairal, Goyal, Bojanowski, and
  Joulin]{Caron2020UnsupervisedLO}
Mathilde Caron, Ishan Misra, J.~Mairal, Priya Goyal, Piotr Bojanowski, and
  Armand Joulin.
\newblock Unsupervised learning of visual features by contrasting cluster
  assignments.
\newblock \emph{ArXiv}, abs/2006.09882, 2020.

\bibitem[Chen et~al.(2020)Chen, Kornblith, Norouzi, and Hinton]{Chen2020ASF}
Ting Chen, Simon Kornblith, Mohammad Norouzi, and Geoffrey~E. Hinton.
\newblock A simple framework for contrastive learning of visual
  representations.
\newblock \emph{ArXiv}, abs/2002.05709, 2020.

\bibitem[Cimpoi et~al.(2014)Cimpoi, Maji, Kokkinos, Mohamed, and
  Vedaldi]{Cimpoi2014DescribingTI}
M.~Cimpoi, Subhransu Maji, I.~Kokkinos, S.~Mohamed, and A.~Vedaldi.
\newblock Describing textures in the wild.
\newblock \emph{2014 IEEE Conference on Computer Vision and Pattern
  Recognition}, pages 3606--3613, 2014.

\bibitem[Cole et~al.(2021)Cole, Yang, Wilber, Aodha, and
  Belongie]{cole2021does}
Elijah Cole, Xuan Yang, Kimberly Wilber, Oisin~Mac Aodha, and Serge Belongie.
\newblock When does contrastive visual representation learning work?, 2021.

\bibitem[D'Amour et~al.(2020)D'Amour, Heller, Moldovan, Adlam, Alipanahi,
  Beutel, Chen, Deaton, Eisenstein, Hoffman, et~al.]{d2020underspecification}
Alexander D'Amour, Katherine Heller, Dan Moldovan, Ben Adlam, Babak Alipanahi,
  Alex Beutel, Christina Chen, Jonathan Deaton, Jacob Eisenstein, Matthew~D
  Hoffman, et~al.
\newblock Underspecification presents challenges for credibility in modern
  machine learning.
\newblock \emph{arXiv preprint arXiv:2011.03395}, 2020.

\bibitem[Dosovitskiy et~al.(2016)Dosovitskiy, Fischer, Springenberg,
  Riedmiller, and Brox]{Dosovitskiy2016DiscriminativeUF}
A.~Dosovitskiy, P.~Fischer, Jost~Tobias Springenberg, Martin~A. Riedmiller, and
  T.~Brox.
\newblock Discriminative unsupervised feature learning with exemplar
  convolutional neural networks.
\newblock \emph{IEEE Transactions on Pattern Analysis and Machine
  Intelligence}, 38:\penalty0 1734--1747, 2016.

\bibitem[Ericsson et~al.(2020)Ericsson, Gouk, and
  Hospedales]{Ericsson2020HowWD}
Linus Ericsson, Henry Gouk, and Timothy~M. Hospedales.
\newblock How well do self-supervised models transfer?
\newblock \emph{ArXiv}, abs/2011.13377, 2020.

\bibitem[Goyal et~al.(2019)Goyal, Mahajan, Gupta, and
  Misra]{Goyal2019ScalingAB}
Priya Goyal, D.~Mahajan, A.~Gupta, and Ishan Misra.
\newblock Scaling and benchmarking self-supervised visual representation
  learning.
\newblock \emph{2019 IEEE/CVF International Conference on Computer Vision
  (ICCV)}, pages 6390--6399, 2019.

\bibitem[Grill et~al.(2020)Grill, Strub, Altch'e, Tallec, Richemond,
  Buchatskaya, Doersch, Pires, Guo, Azar, Piot, Kavukcuoglu, Munos, and
  Valko]{Grill2020BootstrapYO}
Jean-Bastien Grill, Florian Strub, Florent Altch'e, C.~Tallec, Pierre~H.
  Richemond, Elena Buchatskaya, Carl Doersch, B.~A. Pires, Zhaohan~Daniel Guo,
  M.~G. Azar, Bilal Piot, K.~Kavukcuoglu, R.~Munos, and Michal Valko.
\newblock Bootstrap your own latent: A new approach to self-supervised
  learning.
\newblock \emph{ArXiv}, abs/2006.07733, 2020.

\bibitem[Gutmann and Hyv{\"a}rinen(2010)]{Gutmann2010NoisecontrastiveEA}
M.~Gutmann and A.~Hyv{\"a}rinen.
\newblock Noise-contrastive estimation: A new estimation principle for
  unnormalized statistical models.
\newblock In \emph{AISTATS}, 2010.

\bibitem[He et~al.(2020)He, Fan, Wu, Xie, and Girshick]{He2020MomentumCF}
Kaiming He, Haoqi Fan, Yuxin Wu, Saining Xie, and Ross~B. Girshick.
\newblock Momentum contrast for unsupervised visual representation learning.
\newblock \emph{2020 IEEE/CVF Conference on Computer Vision and Pattern
  Recognition (CVPR)}, pages 9726--9735, 2020.

\bibitem[H{\'e}naff et~al.(2020)H{\'e}naff, Srinivas, Fauw, Razavi, Doersch,
  Eslami, and van~den Oord]{Hnaff2020DataEfficientIR}
Olivier~J. H{\'e}naff, A.~Srinivas, J.~Fauw, Ali Razavi, Carl Doersch,
  S.~Eslami, and A{\"a}ron van~den Oord.
\newblock Data-efficient image recognition with contrastive predictive coding.
\newblock \emph{ArXiv}, abs/1905.09272, 2020.

\bibitem[Hendrycks and Dietterich(2019)]{Hendrycks2019BenchmarkingNN}
Dan Hendrycks and Thomas~G. Dietterich.
\newblock Benchmarking neural network robustness to common corruptions and
  perturbations.
\newblock \emph{ArXiv}, abs/1903.12261, 2019.

\bibitem[Horn et~al.(2021)Horn, Cole, Beery, Wilber, Belongie, and
  Aodha]{vanhorn2021benchmarking}
Grant~Van Horn, Elijah Cole, Sara Beery, Kimberly Wilber, Serge Belongie, and
  Oisin~Mac Aodha.
\newblock Benchmarking representation learning for natural world image
  collections, 2021.

\bibitem[Houben et~al.(2013)Houben, Stallkamp, Salmen, Schlipsing, and
  Igel]{Houben2013DetectionOT}
Sebastian Houben, J.~Stallkamp, J.~Salmen, Marc Schlipsing, and C.~Igel.
\newblock Detection of traffic signs in real-world images: The german traffic
  sign detection benchmark.
\newblock \emph{The 2013 International Joint Conference on Neural Networks
  (IJCNN)}, pages 1--8, 2013.

\bibitem[Kolesnikov et~al.(2019)Kolesnikov, Zhai, and
  Beyer]{Kolesnikov2019RevisitingSV}
Alexander Kolesnikov, Xiaohua Zhai, and Lucas Beyer.
\newblock Revisiting self-supervised visual representation learning.
\newblock \emph{2019 IEEE/CVF Conference on Computer Vision and Pattern
  Recognition (CVPR)}, pages 1920--1929, 2019.

\bibitem[Kornblith et~al.(2019)Kornblith, Shlens, and Le]{Kornblith2019DoBI}
Simon Kornblith, Jonathon Shlens, and Quoc~V. Le.
\newblock Do better imagenet models transfer better?
\newblock \emph{2019 IEEE/CVF Conference on Computer Vision and Pattern
  Recognition (CVPR)}, pages 2656--2666, 2019.

\bibitem[Kotar et~al.(2021)Kotar, Ilharco, Schmidt, Ehsani, and
  Mottaghi]{kotar2021contrasting}
Klemen Kotar, Gabriel Ilharco, Ludwig Schmidt, Kiana Ehsani, and Roozbeh
  Mottaghi.
\newblock Contrasting contrastive self-supervised representation learning
  pipelines, 2021.

\bibitem[LeCun et~al.(1998)LeCun, Bottou, Bengio, and
  Haffner]{LeCun1998GradientbasedLA}
Y.~LeCun, L.~Bottou, Yoshua Bengio, and P.~Haffner.
\newblock Gradient-based learning applied to document recognition.
\newblock 1998.

\bibitem[LeCun et~al.(2004)LeCun, Huang, and Bottou]{LeCun2004LearningMF}
Y.~LeCun, F.~Huang, and L.~Bottou.
\newblock Learning methods for generic object recognition with invariance to
  pose and lighting.
\newblock \emph{Proceedings of the 2004 IEEE Computer Society Conference on
  Computer Vision and Pattern Recognition, 2004. CVPR 2004.}, 2:\penalty0
  II--104 Vol.2, 2004.

\bibitem[Maji et~al.(2013)Maji, Rahtu, Kannala, Blaschko, and
  Vedaldi]{Maji2013FineGrainedVC}
Subhransu Maji, Esa Rahtu, Juho Kannala, Matthew~B. Blaschko, and A.~Vedaldi.
\newblock Fine-grained visual classification of aircraft.
\newblock \emph{ArXiv}, abs/1306.5151, 2013.

\bibitem[Misra and Maaten(2020)]{Misra2020SelfSupervisedLO}
Ishan Misra and L.~V.~D. Maaten.
\newblock Self-supervised learning of pretext-invariant representations.
\newblock \emph{2020 IEEE/CVF Conference on Computer Vision and Pattern
  Recognition (CVPR)}, pages 6706--6716, 2020.

\bibitem[Nilsback and Zisserman(2008)]{Nilsback2008AutomatedFC}
Maria-Elena Nilsback and Andrew Zisserman.
\newblock Automated flower classification over a large number of classes.
\newblock \emph{2008 Sixth Indian Conference on Computer Vision, Graphics \&
  Image Processing}, pages 722--729, 2008.

\bibitem[Purushwalkam and Gupta(2020)]{Purushwalkam2020DemystifyingCS}
Senthil Purushwalkam and Abhinav Gupta.
\newblock Demystifying contrastive self-supervised learning: Invariances,
  augmentations and dataset biases.
\newblock \emph{ArXiv}, abs/2007.13916, 2020.

\bibitem[Russakovsky et~al.(2015)Russakovsky, Deng, Su, Krause, Satheesh, Ma,
  Huang, Karpathy, Khosla, Bernstein, Berg, and
  Fei-Fei]{Russakovsky2015ImageNetLS}
Olga Russakovsky, J.~Deng, Hao Su, J.~Krause, S.~Satheesh, S.~Ma, Zhiheng
  Huang, A.~Karpathy, A.~Khosla, Michael~S. Bernstein, A.~Berg, and Li~Fei-Fei.
\newblock Imagenet large scale visual recognition challenge.
\newblock \emph{International Journal of Computer Vision}, 115:\penalty0
  211--252, 2015.

\bibitem[Sagawa et~al.(2019)Sagawa, Koh, Hashimoto, and
  Liang]{Sagawa2019DistributionallyRN}
Shiori Sagawa, Pang~Wei Koh, T.~Hashimoto, and Percy Liang.
\newblock Distributionally robust neural networks for group shifts: On the
  importance of regularization for worst-case generalization.
\newblock \emph{ArXiv}, abs/1911.08731, 2019.

\bibitem[Tian et~al.(2020)Tian, Krishnan, and Isola]{Tian2020ContrastiveMC}
Yonglong Tian, Dilip Krishnan, and Phillip Isola.
\newblock Contrastive multiview coding.
\newblock In \emph{ECCV}, 2020.

\bibitem[Wah et~al.(2011)Wah, Branson, Welinder, Perona, and
  Belongie]{Wah2011TheCB}
C.~Wah, Steve Branson, P.~Welinder, P.~Perona, and Serge~J. Belongie.
\newblock The caltech-ucsd birds-200-2011 dataset.
\newblock 2011.

\bibitem[Wu et~al.(2018)Wu, Xiong, Yu, and Lin]{Wu2018UnsupervisedFL}
Zhirong Wu, Yuanjun Xiong, S.~Yu, and D.~Lin.
\newblock Unsupervised feature learning via non-parametric instance-level
  discrimination.
\newblock \emph{ArXiv}, abs/1805.01978, 2018.

\bibitem[Xiao et~al.(2017)Xiao, Rasul, and Vollgraf]{Xiao2017FashionMNISTAN}
H.~Xiao, K.~Rasul, and Roland Vollgraf.
\newblock Fashion-mnist: a novel image dataset for benchmarking machine
  learning algorithms.
\newblock \emph{ArXiv}, abs/1708.07747, 2017.

\bibitem[Yang et~al.(2020)Yang, He, Liang, Yang, Zhang, and
  Xie]{Yang2020TransferLO}
Xingyi Yang, Xuehai He, Yuxiao Liang, Yue Yang, Shanghang Zhang, and P.~Xie.
\newblock Transfer learning or self-supervised learning? a tale of two
  pretraining paradigms.
\newblock \emph{ArXiv}, abs/2007.04234, 2020.

\bibitem[Zbontar et~al.(2021)Zbontar, Jing, Misra, LeCun, and
  Deny]{zbontar2021barlow}
Jure Zbontar, Li~Jing, Ishan Misra, Yann LeCun, and St{\'e}phane Deny.
\newblock Barlow twins: Self-supervised learning via redundancy reduction.
\newblock \emph{arXiv preprint arXiv:2103.03230}, 2021.

\bibitem[Zhai et~al.(2019)Zhai, Puigcerver, Kolesnikov, Ruyssen, Riquelme,
  Lucic, Djolonga, Pinto, Neumann, Dosovitskiy, Beyer, Bachem, Tschannen,
  Michalski, Bousquet, Gelly, and Houlsby]{Zhai2019ALS}
Xiaohua Zhai, J.~Puigcerver, A.~Kolesnikov, P.~Ruyssen, Carlos Riquelme, Mario
  Lucic, Josip Djolonga, Andr{\'e}~Susano Pinto, Maxim Neumann, A.~Dosovitskiy,
  Lucas Beyer, Olivier Bachem, M.~Tschannen, Marcin Michalski, O.~Bousquet,
  S.~Gelly, and N.~Houlsby.
\newblock A large-scale study of representation learning with the visual task
  adaptation benchmark.
\newblock \emph{arXiv: Computer Vision and Pattern Recognition}, 2019.

\bibitem[Zhao et~al.(2020)Zhao, Wu, Lau, and Lin]{Zhao2020WhatMI}
Nanxuan Zhao, Zhirong Wu, Rynson W.~H. Lau, and Stephen Lin.
\newblock What makes instance discrimination good for transfer learning?
\newblock \emph{ArXiv}, abs/2006.06606, 2020.

\end{thebibliography}

\appendix

\section{Appendix}

\subsection{Details of Datasets Used}
In Table \ref{fig:datasets}, we show the details of the 10 datasets used in this study, including the focus, number of classes, and the size of the training set.

\begin{figure}[h]
\centering
\begin{subfigure}{.48\textwidth}
    \centering
\includegraphics[width=0.8\linewidth]{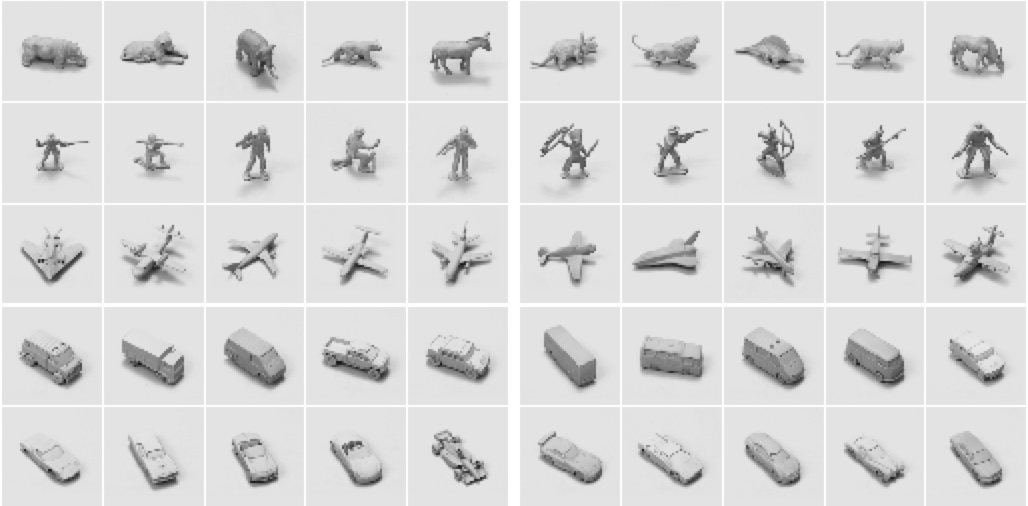}
    \caption{\textbf{NORB.} Figure adapted from \citet{LeCun2004LearningMF}.}
    \label{fig:norbImg}
\end{subfigure}
\hfill
\begin{subfigure}{.48\textwidth}
    \centering
\includegraphics[width=1.0\linewidth]{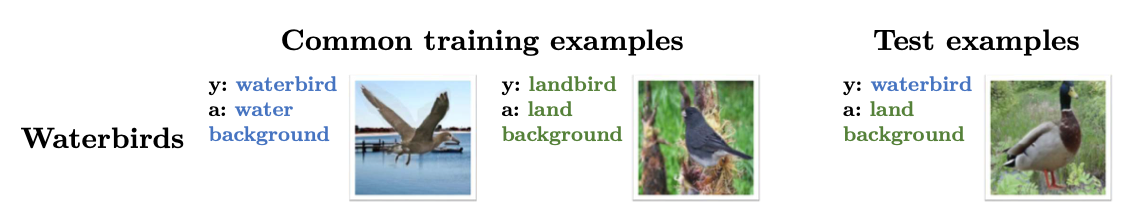}
    \caption{\textbf{Waterbirds.} In Waterbirds, the training examples are constructed such that waterbirds are typically shown on water backgrounds, while landbirds are typically shown on land backgrounds. Testing, however, is conducted on a split of the data where the foreground is independent of the background. Figure adapted from \citet{Sagawa2019DistributionallyRN}.}
    \label{fig:waterbirdsImg}
\end{subfigure}
\\
\vskip 0.2in
\begin{subfigure}{.48\textwidth}
    \centering
\includegraphics[width=0.8\linewidth]{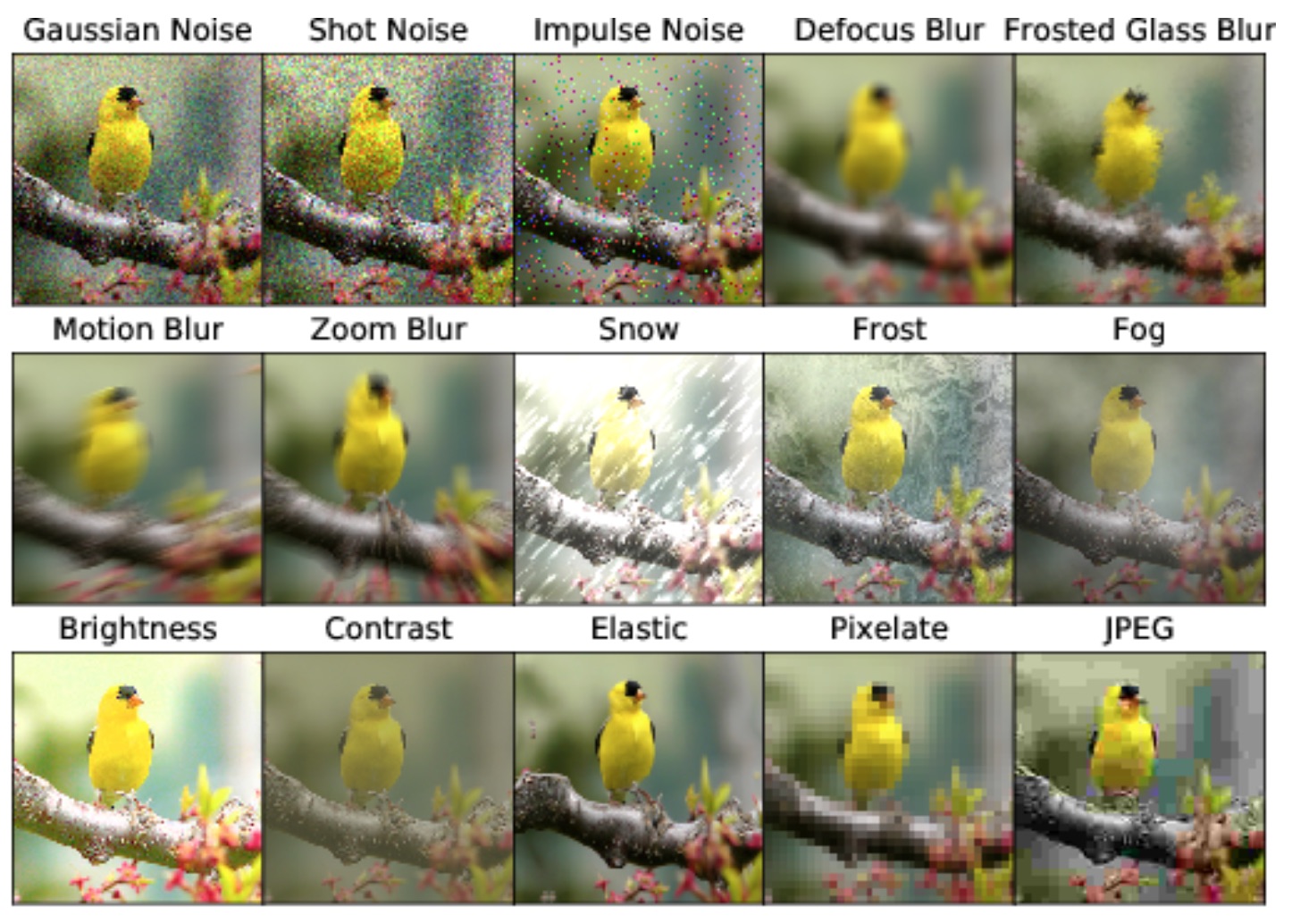}
    \caption{\textbf{ImageNet-C.} Figure adapted from \citet{Hendrycks2019BenchmarkingNN}.}
    \label{fig:ImageNetC_Img}
\end{subfigure}
\caption{\textbf{Example images from datasets used in Section 3.2.2.}}
\label{fig:DatasetExamples}
\end{figure}

Example images from the datasets used in Section 3.2.2 are shown in Figure \ref{fig:DatasetExamples}. Each represents a different way of examining the object-centricity of a model's representations.

\begin{table}
\caption{\textbf{Details of datasets used.}}
\vskip 0.15in
\begin{center}
\begin{small}
\begin{sc}
\begin{tabular}{lcccr}
\toprule
Dataset &  Focus &  Classes &  Train Size \\
\midrule
      VGG Flower &                  flowers &                102 &                   6507 \\
     Traffic Sign &                      road signs &                 43 &                    31367 \\
            MNIST &              digits &                 10 &                  60000 \\
    Fashion MNIST &                         apparel &                 10 &                  60000 \\
              DTD &                        textures &                 47 &                   3760 \\
         CU Birds &                    birds &                200 &                  5994 \\
         Aircraft &         aircrafts &                100 &  3334 \\
         ImageNet &                   diverse &               1000 &                1281167 \\
    Norb-Object &                            toys &                  6 &                  48600 \\
 Norb-Elevation &                      elevations &                  9 &                  48600 \\
  Norb-Lighting &             lighting &                  6 &                  48600 \\
  Norb-Azimuth &                 rotations &                 18 &                  48600 \\
      Waterbirds &  birds &                  2 &                   4795 \\
\bottomrule
\end{tabular}
\end{sc}
\end{small}
\label{fig:datasets}
\end{center}
\vskip -0.1in
\end{table}

\subsection{Transfer Accuracies on Diverse Tasks}
\begin{figure}[h]
\vskip 0pt
\begin{subfigure}{.48\textwidth}
    \centering
\includegraphics[width=0.8\linewidth]{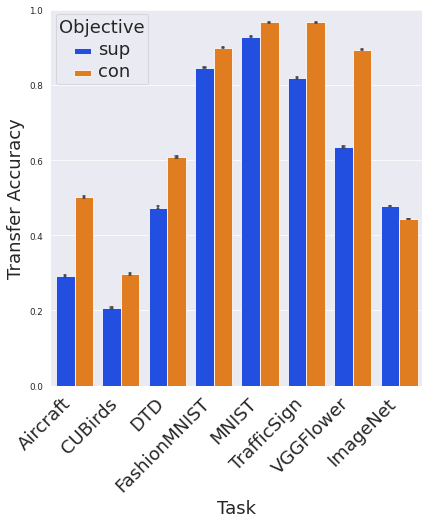}
    \caption{\textbf{Comparison of transfer accuracies achieved by supervised and contrastive pretraining across 8 diverse image classification transfer tasks.} Both models were pretrained on ImageNet for 200 epochs. Contrastive accuracy was higher than supervised accuracy on all tasks except ImageNet. Results on all tasks represent the average of three independent runs, with error bars representing the standard deviation.}
    \label{fig:transfer}
\end{subfigure}
\hfill
\begin{subfigure}{.48\textwidth}
    \centering
\includegraphics[width=1.0\linewidth]{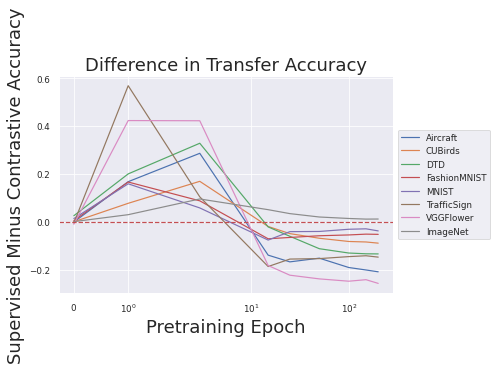}
    \caption{\textbf{Difference in transfer accuracy (supervised minus contrastive) on eight image classification tasks.} The dashed red line indicates when contrastive and supervised accuracies match, and we see that every task except ImageNet crosses the dashed line from positive to negative---indicating that contrastive accuracy overtakes supervised accuracy---at or before epoch 15 of pretraining.}
    \label{fig:difference}
\end{subfigure}
\caption{\textbf{Transfer accuracies on diverse tasks.}}
\label{fig:transfer+difference}
\end{figure}

In Table \ref{fig:comparison}, we compare the transfer accuracies on 8 diverse tasks after 200 epochs of pretraining. We visualize this comparison in Figure \ref{fig:transfer}, and we explore the difference in transfer accuracy between the supervised and contrastive models over the course of pretraining in Figure \ref{fig:difference}.

\begin{table}[t]
\caption{\textbf{Comparison of transfer accuracies on diverse tasks.} After 200 epochs, the contrastive model achieves higher transfer accuracy for all tasks except ImageNet, which was used to pretrain the supervised model. Values after $\pm$ are standard deviations.}
\vskip 0.15in
\begin{center}
\begin{small}
\begin{sc}
\begin{tabular}{lcccr}
\toprule
Task & Supervised & Contrastive \\
\midrule
Aircraft & 29.1 $\pm$ 0.3 & \textbf{50.0} $\pm$ 0.3  \\
CUBirds & 20.7 $\pm$ 0.3 & \textbf{29.7} $\pm$ 0.2  \\
FashionMNIST & 84.6 $\pm$ 0.1 & \textbf{89.9} $\pm$ 0.1  \\
DTD & 47.4 $\pm$ 0.3 & \textbf{60.8} $\pm$ 0.2 \\
TrafficSign & 81.8 $\pm$ 0.2 & \textbf{96.6} $\pm$ 0.1 \\
MNIST & 92.8 $\pm$ 0.1 & \textbf{96.7} $\pm$ 0.1 \\
VGGFlower & 63.6 $\pm$ 0.2 & \textbf{89.4} $\pm$ 0.1 \\
ImageNet & \textbf{47.8} $\pm$ 0.0 & 44.4 $\pm$ 0.0\\
\bottomrule
\end{tabular}
\end{sc}
\end{small}
\label{fig:comparison}
\end{center}
\vskip -0.25in
\end{table}

\end{document}